\documentclass[runningheads]{llncs}
\usepackage[T1]{fontenc}
\usepackage{graphicx}
\usepackage{bm}
\usepackage{amsmath, amssymb}
\usepackage{algpseudocode}
\usepackage{multirow}
\usepackage{booktabs}
\usepackage{marvosym}
\usepackage{lipsum}
\usepackage{listings}
\usepackage[colorlinks=true,linkcolor=blue,citecolor=blue,urlcolor=blue,]{hyperref}
\begin{document}
\title{One-Shot Medical Video Object Segmentation via Temporal Contrastive Memory Networks}
\titlerunning{Temporal Contrastive Memory Networks}
%
\author{Yaxiong Chen\inst{1,2} \and 
Junjian Hu\inst{1}\thanks{Work done during an internship at MedAI Technology (Wuxi) Co. Ltd.} \and
Chunlei Li\inst{3} \and
Zixuan Zheng\inst{3} \and
Jingliang Hu\inst{3} \and
Yilei Shi\inst{3} \and
Shengwu Xiong\inst{1,2} \and
Xiao Xiang Zhu\inst{4} \and
Lichao Mou\inst{3}\textsuperscript{(\Letter)}}


\authorrunning{Y. Chen et al.}
\institute{Wuhan University of Technology, Wuhan, China \and Shanghai Artificial Intelligence Laboratory, Shanghai, China \and MedAI Technology (Wuxi) Co. Ltd., Wuxi, China\\\email{lichao.mou@medimagingai.com} \and Technical University of Munich, Munich, Germany}

\maketitle              

\begin{abstract}
Video object segmentation is crucial for the efficient analysis of complex medical video data, yet it faces significant challenges in data availability and annotation. We introduce the task of one-shot medical video object segmentation, which requires separating foreground and background pixels throughout a video given only the mask annotation of the first frame. To address this problem, we propose a temporal contrastive memory network comprising image and mask encoders to learn feature representations, a temporal contrastive memory bank that aligns embeddings from adjacent frames while pushing apart distant ones to explicitly model inter-frame relationships and stores these features, and a decoder that fuses encoded image features and memory readouts for segmentation. We also collect a diverse, multi-source medical video dataset spanning various modalities and anatomies to benchmark this task. Extensive experiments demonstrate state-of-the-art performance in segmenting both seen and unseen structures from a single exemplar, showing ability to generalize from scarce labels. This highlights the potential to alleviate annotation burdens for medical video analysis. Code is available at \url{https://github.com/MedAITech/TCMN}.


\end{abstract}
\section{Introduction}
Video object segmentation is pivotal for efficiently processing complex medical video data and empowering accurate analysis. Precise segmentation of varying tissue regions and lesions facilitates early and accurate diagnosis of lesion characteristics. Furthermore, tracking lesions during surgical procedures aids surgeons in localization, enabling more effective treatment. 
\par
Yet medical video object segmentation faces key data and annotation challenges. While pioneering works have explored lesion and nerve segmentation from ultrasound and endoscopy~\cite{FLA23,FU23}, acquiring sizable labeled medical video data places high burden on domain experts. Semi-supervised approaches~\cite{SPN23,TCE22} reduce annotations but still rely on class-specific features, limiting generalization to unseen classes.
\par
To surmount the above limitations, we introduce one-shot medical video object segmentation---accurately separating foreground and background throughout a video sequence given a first frame annotation. Our goal is not only to segment known objects, but, crucially, to generalize to segment new classes without any fine-tuning or re-training.
\par
The essence of video object segmentation lies in exploiting dynamic cues within video frames to propagate mask information across the sequence. In the one-shot setting, a mask annotation is provided for the first frame, offering a valuable foreground localization cue to leverage for subsequent unlabeled frames. Recent video object segmentation methods employ memory networks to model inter-frame dependencies. \cite{STM19} pioneers the integration of memory networks into this task. \cite{STCN21} refines affinity calculations to improve propagation. \cite{RDE22} compresses stored representations with a recurrent embedding module. By computing spatiotemporal attention between query pixels and past frames, these memory-augmented models determine pixel-wise foreground likelihoods.
\par
Prevailing memory-based approaches primarily emphasize feature storage and retrieval, without adequately modeling the temporal context between frames. This may overlook crucial inter-frame dependencies that encode video dynamics. We propose to incorporate contrastive learning into the memory framework to explicitly optimize the storage and use of temporal information. Our core insight is that embeddings of frames belonging to the same semantic class should exhibit proximity that decays with increasing temporal distance; that is, the learned embeddings have greater proximity within the memory when from adjacent timesteps, while displaying more divergence for distant timesteps. To realize this idea, we propose a temporal contrastive memory network for video object segmentation. Our model comprises four main components---an image encoder, a mask encoder, a temporal contrastive memory bank, and a decoder. The image and mask encoders learn image and mask representations of each frame, respectively. Both representations are stored in the memory bank. To model intricate video dynamics, we devise a temporal contrastive loss within the memory bank which aligns feature representations of adjacent frames in the memory while pushing apart those from distant frames. This enhances inter-frame relationship modeling. The decoder fuses encoded image features and memory readouts to predict segmentations.
\par
To benchmark one-shot video object segmentation in medical imaging, we introduce a multi-modal, multi-organ dataset compiled from colonoscopy and cardiac ultrasound videos across four sources: ASU-Mayo~\cite{ASU}, CVC-ClinicDB~\cite{CVC600}, HMC-QU~\cite{HMC}, and CAMUS~\cite{CAMUS}. Our key contributions are:
\begin{itemize}
    \item We formulate and investigate the task of one-shot segmentation in medical videos.
    \item We assemble a diverse medical video dataset, establishing a benchmark for this problem.
    \item We propose a temporal contrastive memory network. Extensive experiments demonstrate state-of-the-art performance in segmenting both seen and unseen anatomical structures or lesions from a single exemplar.
\end{itemize}

\section{Methodology}

\begin{figure}[!th]
  \centering
  \includegraphics[width=\textwidth]{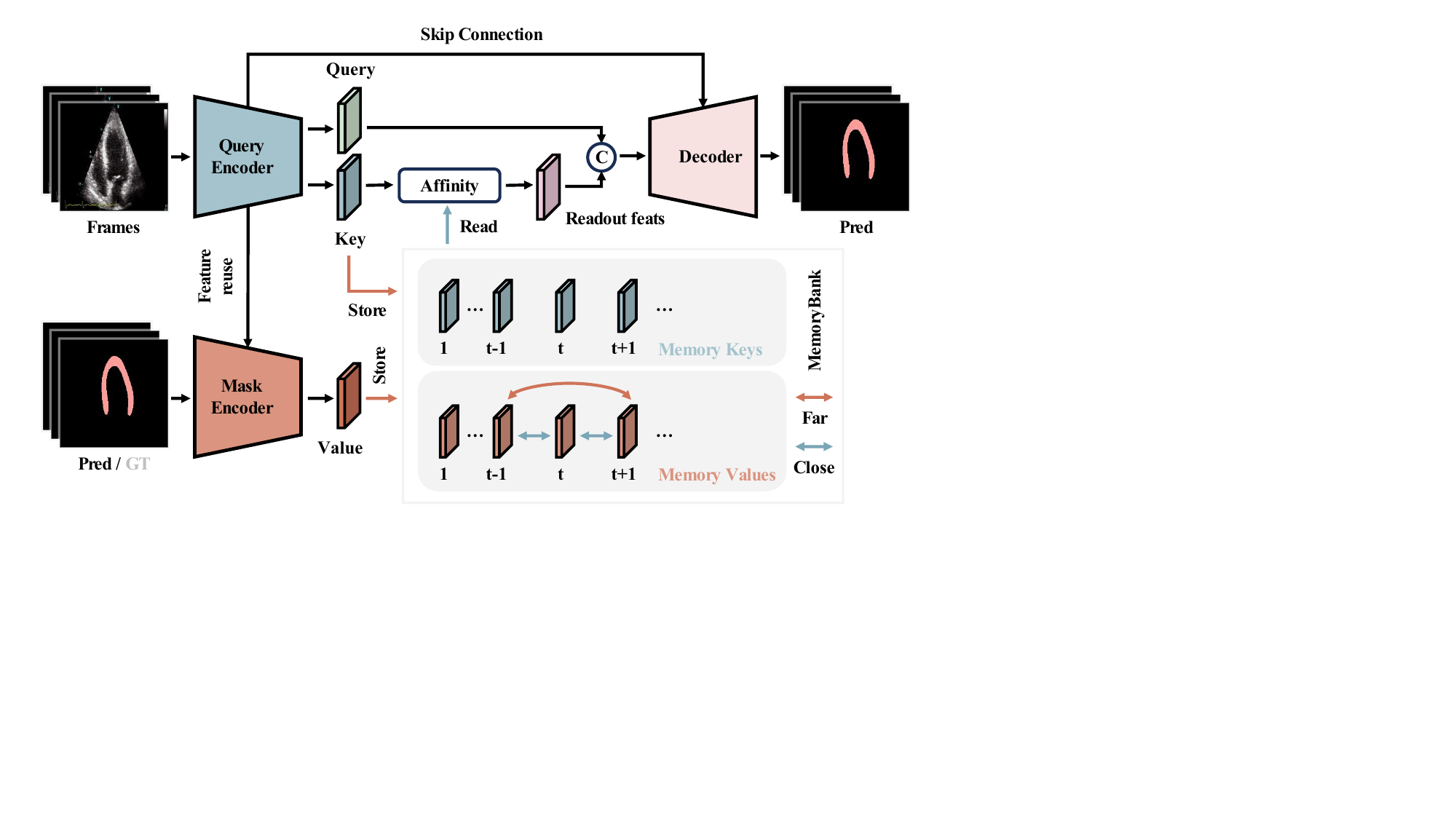}
  \caption{Overview of our proposed model for one-shot medical video object segmentation. Our model consists of four main components: an image encoder, a mask encoder, a temporal contrastive memory bank, and a decoder. We feed the first frame and its corresponding mask annotation into the image encoder and mask encoder, respectively, to generate key and value feature maps. These are stored in the memory bank as initial memories. For subsequent frames, we input them into the image encoder to generate new key features and utilize the memory bank to produce segmentations. Concurrently, we feed the predicted segmentations to the mask encoder to generate new value features, which, along with the newly generated key features, are stored in the memory bank as updated memories. Our designed memory bank optimizes the memory information through temporal contrastive learning. This process is repeated until the entire video is segmented.}\label{fig1}
\end{figure}

Our objective is to segment the remaining frames of a video sequence, given the annotation of the initial frame. As depicted in Fig.~\ref{fig1}, the proposed framework comprises four key components: a query encoder, a mask encoder, a memory bank, and a mask decoder. Notably, to leverage relevant historical frame information during the segmentation process of subsequent frames, we construct a continuously updated memory bank. This memory bank serves as a repository for features extracted from both the query encoder and the mask encoder, facilitating the propagation of temporal cues. In the following sections, we provide a comprehensive elucidation of each component.

\subsection{Query Encoder}
The query encoder leverages a modified ResNet-50~\cite{ResNet} architecture, where the classification head and the final convolutional stage are omitted. Subsequently, we extend the network by appending two independent projection heads. Consequently, the query encoder takes single video frames as input and generates pairs of query and key feature maps as outputs. The latter are cached in the memory bank, serving as new memory keys to facilitate temporal reasoning.

\subsection{Temporal Contrastive Mask Encoding}
The mask encoder in our framework employs a customized ResNet-18~\cite{ResNet} backbone, where the classification head and the final convolutional stage are removed. Instead, we append two additional convolutional layers. The input to the mask encoder is object masks. Furthermore, we reuse outputs of the query encoder (prior to the projection heads) as an auxiliary input to the mask encoder. Specifically, object masks are initially encoded by the backbone and subsequently concatenated with the reused features. These joint representations are fed into the convolutional layers to produce feature maps, denoted as value features. The value features are stored in the memory bank alongside the corresponding key features generated by the query encoder. Thus, we efficiently establish two distinct sets within the memory bank: one allocated for key features, and the other for value features.
\par
The quality of memory values in the memory bank has a direct impact on the segmentation of subsequent frames. Therefore, we further incorporate temporal contrastive learning into our framework. Specifically, for the value set $\mathcal{V}$, our hypothesis is that adjacent memory values in time exhibit stronger semantic similarity. This is because objects in consecutive video frames typically have similar shapes and appear in similar backgrounds, hence their encoded features should have similar representations. On the contrary, for distant memory values in time, we expect that their feature representations are relatively dissimilar. This design enables our model to effectively utilize temporal information for memory.
\par

To utilize the information of previous frames to assist in the segmentation of the current frame, we employ a simple yet efficient method for retrieving relevant information from the memory bank. Specifically, we use an evaluation model based on negative squared Euclidean distance~\cite{NSED07} to match memory keys in the memory bank with key feature maps of the current frame obtained from the query encoder. This calculation involves performing pixel-level computations on feature maps to generate a similarity matrix. Finally, by the simple matrix multiplication of memory values and softmax-normalized similarity matrix, we obtain readout feature maps for the query frame.

\subsection{Mask Decoder}
The readout features are concatenated with query features. This concatenated representation is then fed into the mask decoder to predict segmentation masks. Specifically, the mask decoder comprises interleaved convolutional and upsampling layers, followed by a softmax operation.

\subsection{Overall Loss}
To enhance the quality of memory values stored in the memory bank, we propose a novel temporal contrastive loss:
\begin{equation}
\mathcal{L}_{\mathrm{tc}}=\sum_{t=1}^{|\mathcal{V}|}\frac{(1-\mathrm{sim}(\bm{v}_{t-1},\bm{v}_{t}))+(1-\mathrm{sim}(\bm{v}_{t},\bm{v}_{t+1}))}{1-\mathrm{sim}(\bm{v}_{t-1},\bm{v}_{t+1})+\epsilon} \,,
\end{equation}
where $\mathrm{sim}(\bm{a},\bm{b})=\bm{a}^\mathrm{T}\bm{b}/\left \|\bm{a}\right \|\left \|\bm{b}\right \|$ denotes the cosine similarity between vectors $\bm{a}$ and $\bm{b}$, and $\bm{v}_{t}$ represents the $t$-th memory value in the memory bank. A small value $\epsilon$ is added to the denominator to prevent division-by-zero errors in practice. We use bootstrapped cross entropy (BCE)~\cite{BCE} as the supervised loss:
\begin{equation}
\mathcal{L}_{\mathrm{bce}}=\frac{1}{|\mathcal{D}|}\sum_{\bm{y}_i\in\mathcal{D}}\{\hat{\bm{y}}_i<\eta\}\mathbf{H}(\bm{y}_i,\hat{\bm{y}}_i) \,,
\end{equation}
where $\mathcal{D}$ is the set of frames in a video, excluding the first frame. $\hat{\bm{y}}_i$ is the predicted segmentation mask of the $i$-th frame, and $\bm{y}_i$ is the corresponding ground truth mask. $\mathbf{H}(\cdot)$ denotes cross entropy loss. We only calculate the loss for pixels with probabilities less than a threshold $\eta$ to prevent over-training on easy samples. The overall loss function of our model is:
\begin{equation}
\mathcal{L}=\alpha\mathcal{L}_{\mathrm{bce}}+\beta\mathcal{L}_{tc} \,,
\end{equation}
where $\alpha$ and $\beta$ are two coefficients balancing the two loss terms.

\section{Experiments}

\subsection{Experimental Settings}
\subsubsection{Datasets}
We conduct experiments on four public medical video datasets. The first is the ASU-Mayo colonoscopy video dataset, containing 10 negative videos from normal subjects and 10 positive videos from patients. We use the positive videos, consisting of 5,402 video frames in total, with 3,799 frames containing polyps, for our experiments. The second dataset is CVC-ClinicDB, comprising 29 colonoscopy videos, totaling 612 frames, all of which contain polyps. The third dataset is HMC-QU, containing 109 apical-4-chamber (A4C) view echocardiography videos with left ventricle wall segmentation masks. The final dataset is CAMUS, containing 500 patient samples, each including apical-2-chamber (A2C) and A4C echocardiography videos with segmentation masks for left ventricle wall, left ventricle, and left atrium. The first three datasets, i.e., ASU-Mayo, CVC-ClinicDB, and HMC-QU, are divided into training and test sets with an 8:2 ratio. The training sets from the three datasets are then combined and used to train models. It is worth noting that, to validate models' generalization capabilities, we use the CAMUS dataset, which contains two classes not present in training videos.

\subsubsection{Evaluation Metrics}

We employ three widely used metrics, Jaccard index $\mathcal{J}$~\cite{Jaccard}, contour accuracy $\mathcal{F}$, and $Dice$, to evaluate the performance of models. $\mathcal{J}$ is defined as the intersection-over-union of an estimated segmentation mask and the corresponding ground truth. 
$\mathcal{F}$ is defined by treating a mask as a set of closed contours and computing a contour-based F-measure. 

\subsubsection{Implementation Details}

Our experiments are conducted on a single 24 GB NVIDIA GeForce RTX 4090 GPU with the Adam optimizer using PyTorch. Following previous practices~\cite{STM19,STCN21,RDE22}, we first pre-train models on static image datasets~\cite{Data1,Data2,Data3} with synthetic deformation and then use real video data for main training. We use randomly cropped $384\times384$ patches and a batch size of 16 during pre-training. A batch size of 8 is adopted during main training. In addition, following a training strategy present in \cite{STM19,STCN21}, we sample 3 temporally ordered frames from a training video by randomly skipping 0-5 frames. During inference, we designate every fifth frame as a memory frame. Given the relatively short duration of video sequences in the dataset, we opt not to impose a fixed limit on the size of memory bank. Instead, the memory bank capacity is determined by available resources on the target inference platform.

\begin{table}[!t]\scriptsize
\caption{Quantitative evaluation of the proposed method and other approaches on the ASU-Mayo, CVC-ClinicDB, and HMC-QU datasets for one-shot medical video object segmentation. ``Mean'' denotes the average performance across the three datasets.}
\label{seen_table}
\begin{center}
\renewcommand\arraystretch{1}
\setlength{\tabcolsep}{2.2pt}
\begin{tabular}{lcccccccccccc}
& \multicolumn{3}{c}{\textbf{ASU-Mayo}} & \multicolumn{3}{c}{\textbf{CVC-ClinicDB}} & \multicolumn{3}{c}{\textbf{HMC-QU}} & \multicolumn{3}{c}{\textbf{Mean}} \\ \midrule[0.85pt]
Methods              & $\mathcal{J}$        & $\mathcal{F}$       & $Dice$    & $\mathcal{J}$         & $\mathcal{F}$         & $Dice$     & $\mathcal{J}$       & $\mathcal{F}$       & $Dice$   & $\mathcal{J}$      & $\mathcal{F}$      & $Dice$   \\ \midrule[0.5pt]
AFB-URR              & 62.23    & 70.02   & 68.98   & 77.22     & 68.07     & 85.64    & 86.59   & 98.91   & 92.76  & 75.34  & 79.00  & 82.79  \\
STCN                 & 68.58    & 76.97   & 77.30   & 82.72     & 77.76     & 89.27    & 89.94   & 99.38   & 94.66  & 80.41  & 84.70  & 87.08  \\
AOT                  & 67.29    & 75.24   & 75.57   & 77.03     & 65.38     & 85.18    & 77.49   & 94.55   & 87.25  & 73.93  & 78.39  & 82.67  \\
JOINT                & 70.08    & 77.71   & 78.79   & 77.27     & 67.07     & 86.01    & 84.36   & 97.65   & 91.44  & 77.24  & 79.00  & 85.41  \\
DeAOT                & 70.68    & 79.79   & 79.56   & 74.24     & 64.89     & 82.89    & 77.11   & 94.55   & 87.00  & 74.01  & 79.86  & 83.15  \\
RPCM                 & 58.49    & 62.72   & 66.15   & 62.07     & 48.95     & 73.73    & 38.13   & 52.47   & 51.05  & 52.90  & 54.71  & 63.64  \\
RDE                  & 61.89    & 69.00   & 70.49   & 77.18     & 70.66     & 85.53    & 87.65   & 98.94   & 93.36  & 75.57  & 79.53  & 83.13  \\
XMem                 & 67.58    & 76.12   & 76.15   & 82.74     & 76.46     & 89.58    & 90.41   & 99.38   & 94.92  & 80.24  & 83.99  & 86.88  \\
Cutie                & 61.89    & 69.32   & 67.99   & 66.74     & 56.58     & 74.88    & 72.21   & 87.99   & 83.61  & 66.74  & 70.96  & 75.49  \\
\midrule[0.5pt]
w/o $\mathcal{L}_{tc}$    & 70.52    & 79.48   & 79.48   & 84.16     & 77.77     & 90.57    & 90.25   & \textbf{99.50}   & 94.83  & 81.64  & 85.58  & 88.30  \\
Ours                 & \textbf{71.34}    & \textbf{80.13}   & \textbf{80.36}   & \textbf{84.82}     & \textbf{80.61}     & \textbf{90.75}    & \textbf{90.50}   & 99.47   & \textbf{94.98}  & \textbf{82.22}  & \textbf{86.74}  & \textbf{88.70}  \\ \midrule[0.85pt]
\end{tabular}
\end{center}
\end{table}

\subsection{Comparison with State-of-the-Art Methods}

To evaluate the performance of our method, we extensively compare it with state-of-the-art approaches, including  STCN~\cite{STCN21}, AOT~\cite{AOT21}, DeAOT~\cite{DeAOT22}, RDE~\cite{RDE22}, AFB-URR~\cite{AFB-URR20}, JOINT~\cite{JOINT21}, RPCM~\cite{RPCM22}, XMem~\cite{XMem22}, and Cutie~\cite{Cutie23}. To ensure a fair and unbiased comparison, we obtain the segmentation results of all competing methods through their publicly available implementations.

\subsubsection{Quantitative Comparison}
For performance on test sets with seen classes, Table~\ref{seen_table} reports numerical results. It can be seen that our method outperforms competing approaches on every dataset. As for testing on unseen classes, experimental results can be found in Table~\ref{unseen_table}, where our method achieves the second-best results on the left ventricular category and the best results on the left atrial category. On average, the proposed model surpasses competitors.

\begin{table}[!t]\scriptsize
\caption{Quantitative evaluation of the proposed method and competing approaches on two unseen classes, left ventricle and left atrium, from the CAMUS dataset.}
\label{unseen_table}
\begin{center}
\renewcommand\arraystretch{1}
\setlength{\tabcolsep}{3.4pt}
\begin{tabular}{lccccccccc}
& \multicolumn{3}{c}{\textbf{Left Ventricle}} & \multicolumn{3}{c}{\textbf{Left Atrium}} &  \multicolumn{3}{c}{\textbf{Mean}} \\ \midrule[0.85pt]
Methods              & $\mathcal{J}$        & $\mathcal{F}$       & $Dice$    & $\mathcal{J}$         & $\mathcal{F}$         & $Dice$     & $\mathcal{J}$       & $\mathcal{F}$       & $Dice$      \\ \midrule[0.5pt]
AFB-URR              & 66.65    & 39.61   & 77.55   & 51.83     & 30.32     & 63.50    & 59.24   & 34.97   & 70.52    \\
STCN                 & 75.98    & 54.72   & 84.91   & 72.46     & 57.24     & 81.12    & 74.22   & 55.98   & 83.02    \\
AOT                  & 79.09    & \textbf{61.02}   & 86.44   & 63.64     & 40.89     & 74.31    & 71.37   & 50.96   & 80.38    \\
JOINT                & 62.22    & 29.65   & 75.00   & 53.87     & 24.66     & 66.03    & 58.04   & 27.16   & 70.52    \\
DeAOT                & 78.73    & 53.05   & 87.75   & 51.60     & 32.10     & 61.12    & 65.17   & 42.57   & 74.43    \\
RPCM                 & 54.89    & 26.44   & 69.01   & 44.44     & 21.53     & 58.09    & 49.67   & 23.98   & 63.55    \\
RDE                  & 20.64    & 26.75   & 29.23   & 21.14     & 27.36     & 29.55    & 20.89   & 27.05   & 29.39    \\
XMem                 & 72.27    & 37.98   & 83.16   & 80.34     & 57.68     & 88.70    & 76.30   & 47.83   & 85.93    \\
Cutie                & \textbf{82.59}    & 56.36   & \textbf{90.31}   & 72.67     & 49.87     & 82.46    & 77.63   & 53.11   & 86.39    \\
\midrule[0.5pt]
w/o $\mathcal{L}_{tc}$    & 80.06    & 56.46   & 84.91   & 73.80     & 57.24     & 81.12    & 76.93   & 55.98   & 83.02    \\
Ours                 & 81.56    & 58.55   & 89.35   & \textbf{83.10}     & \textbf{66.37}     & \textbf{90.33}    & \textbf{82.33}   & \textbf{62.46}   & \textbf{89.84}    \\ \midrule[0.85pt]
\end{tabular}
\end{center}
\end{table}

\begin{figure}[!th]
  \centering
  \includegraphics[width=\textwidth]{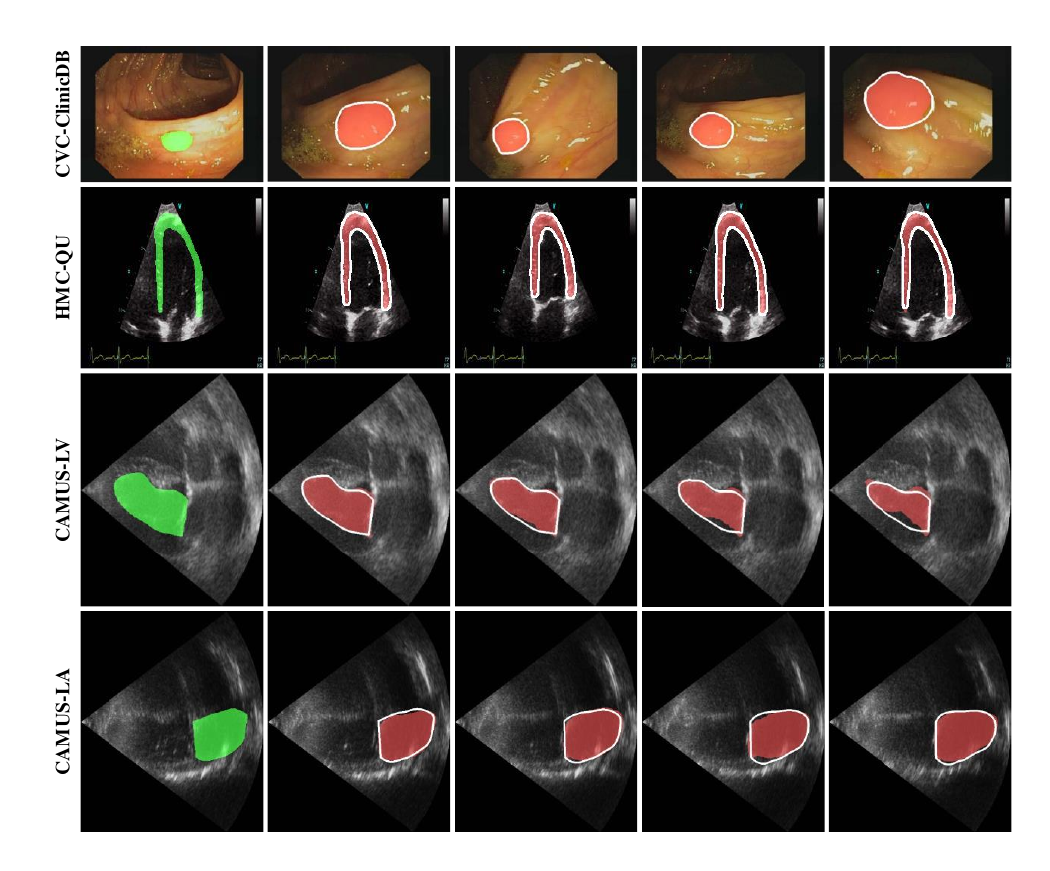}
  \caption{Qualitative results. The first two rows show segmentation results on classes that our model has learned during training, while the last two rows show results on classes that the model has not seen during training. The green masks indicate annotation masks of the first frames within videos. The red masks represent predicted segmentation masks obtained by our model on subsequent frames, and white lines indicate the corresponding ground truth contours for these frames.}\label{fig2}
\end{figure}

\subsubsection{Qualitative Comparison}

Fig.~\ref{fig2} visually presents the qualitative performance of our model on different classes. The first two rows show segmentation results on classes that the model has learned, while the last two rows show results on classes that the model has not seen before. Notably, our model shows good segmentation performance even in these unseen classes, confirming that it learns objectness cues rather than class-specific features.

\subsection{Ablation Study}
To verify the effectiveness of key components of our model, we conduct extensive ablation experiments on four datasets. We take a model without the temporal contrastive loss as the baseline. Upon incorporating this loss, our model witnesses improvements of 0.58\%, 1.16\%, and 0.4\% in average $\mathcal{J}$, $\mathcal{F}$, and $Dice$, respectively, for the seen classes, and notable gains of 5.40\%, 6.48\%, and 6.82\% for the unseen classes.

\section{Conclusion}
In this paper, we introduce the task of one-shot video object segmentation for medical videos and propose a temporal contrastive memory network to address the key challenges associated with limited annotations. We conduct extensive experiments on our diverse, multi-source medical video benchmark, demonstrating that our approach achieves state-of-the-art performance. By leveraging a contrastive memory mechanism, our model can accurately segment both seen and unseen anatomical structures or lesions from only the first frame's annotation. This highlights our method's potential to significantly alleviate laborious annotation efforts required for understanding medical videos. In future work, we plan to explore enhancements to memory banks, focusing on optimizing memory sizes and developing a dynamic selection strategy for memory frames. These improvements aim to increase the applicability of our method across diverse scenarios.

\subsubsection{Prospect of application:}
The proposed method can serve as an interactive, intelligent medical video annotation tool. Users need only provide annotation for the first frame, after which the model generates segmentation masks for subsequent frames, significantly reducing annotation burdens for clinicians. Furthermore, experiments demonstrate the generalizability of this approach.

\begin{credits}
\subsubsection{\ackname} This work is supported in part by the National Key Research and Development Program of China (2022ZD0160604), in part by the Natural Science Foundation of China (62101393/62176194), in part by the High-Performance Computing Platform of YZBSTCACC, and in part by MindSpore (\url{https://www.mindspore.cn}), a new deep learning framework.

\subsubsection{\discintname}
The authors have no competing interests to declare that are relevant to the content of this paper.
\end{credits}

%
%

\end{document}